\documentclass{article}

\usepackage{arxiv}

\usepackage[utf8]{inputenc} 
\usepackage[T1]{fontenc}    
\usepackage{hyperref}       
\usepackage{url}            
\usepackage{booktabs}       
\usepackage{amsfonts}       
\usepackage{nicefrac}       
\usepackage{microtype}      
\usepackage[pdftex]{graphicx}
\usepackage{float}
\usepackage{listings}
\usepackage{booktabs}
\usepackage{amsmath}

\title{Compositional pre-training for neural semantic parsing}

\author{
  Amir Ziai\\
  Department of Computer Science\\
  Stanford University\\
  \texttt{amirziai@stanford.edu}
}

\begin{document}
\maketitle

\begin{abstract}
Semantic parsing is the process of translating natural language utterances into logical forms, which has many important applications such as question answering and instruction following. Sequence-to-sequence models have been very successful across many NLP tasks. However, a lack of task-specific prior knowledge can be detrimental to the performance of these models. Prior work has used frameworks for inducing grammars over the training examples, which capture conditional independence properties that the model can leverage. Inspired by the recent success stories such as BERT we set out to extend this augmentation framework into two stages. The first stage is to pre-train using a corpus of augmented examples in an unsupervised manner. The second stage is to fine-tune to a domain-specific task. In addition, since the pre-training stage is separate from the training on the main task we also expand the universe of possible augmentations without causing catastrophic inference. We also propose a novel data augmentation strategy that interchanges tokens that co-occur in similar contexts to produce new training pairs. We demonstrate that the proposed two-stage framework is beneficial for improving the parsing accuracy in a standard dataset called GeoQuery for the task of generating logical forms from a set of questions about the US geography.
\end{abstract}


\section{Introduction}
Semantic parsing is the task of converting natural language into machine-executable logical forms. Examples of this parsing include asking questions that are then converted to queries against a database, generating code from natural language, converting natural language instructions to an instruction set that can be followed by a system, and even converting natural language into Python \cite{yin2017syntactic}. These logical forms can be captured using notions of formal semantics in linguistic such as $\lambda$-calculus and a more compact version called lambda dependency-based compositional semantics or $\lambda$-DCS \cite{liang2013lambda}.

Traditionally this task has been tackled by a combination of heuristics and search to build up parsers from large datasets of question-answer pairs \cite{berant2013semantic} or from text that is paired with knowledge base information \cite{berant2014semantic}. However, we the advent of the sequence-to-sequence \cite{sutskever2014sequence} architecture, the majority of the research has shifted towards using this framework.

Many sequence-to-sequence use cases involve converting a sequence of natural language into another sequence of natural language. Semantic parsing is different in that the decoded sequence need to be constrained by what would constitute a valid logical form. This additional challenge adds extra complexity to semantic parsing systems. Similar to more conventional sequence-to-sequence tasks semantic parsing also suffers from the problem that one source sentence can have multiple valid logical forms which introduces a wrinkle in evaluation.

\section{Related Work}
\label{sec:headings}

The approach used in this work is a continuation of the work by Jia et al. \cite{jia2016data} where the authors proposed a sequence-to-sequence model with an attention-based copying mechanism. This supervised approach leverages the flexibility of of the encoder-decoder architecture and the authors demonstrate that the model can learn very accurate parsers across three standard semantic parsing datasets. The augmentation strategy used in this work allows for injecting prior knowledge which improve the generalization power of the model.

One disadvantage of this approach is that the decoder outputs are considered unstructured and can lead to invalid logical forms. Krishnamurthy et al. propose to overcome this problem by imposing a grammar on the decoder that only generates well-typed logical forms \cite{krishnamurthy2017neural}. However, this approach increases the complexity of the system, which was unwarranted in our experiments and most of the results produced by the model in our experiments were valid logical forms (with sufficient training).

Moreover, building annotated semantic parsing datasets is highly labor-intensive and parsers built for one domain do not necessarily transfer across domains. Fan et al. propose a multi-task setup and demonstrate that training using this setup can improve the accuracy in domains with smaller labeled datasets \cite{fan2017transfer}. This approach is aligned with one of the main contributions in our research. Our proposed framework allows for pre-training the model in an unsupervied manner with data from multiple tasks that enables transfer learning.

\section{Approach}
A sequence-to-sequence model with an attention-based copying mechanism is used to learn logical forms from natural language utterances. Moreover, a novel data augmentation framework is used for injecting prior knowledge by inducing a synchronous context-free grammar. This novel framework, first proposed in \cite{jia2016data}, is called data recombination.

\begin{figure}[H]
    \centering
        \includegraphics[width=0.8\linewidth]{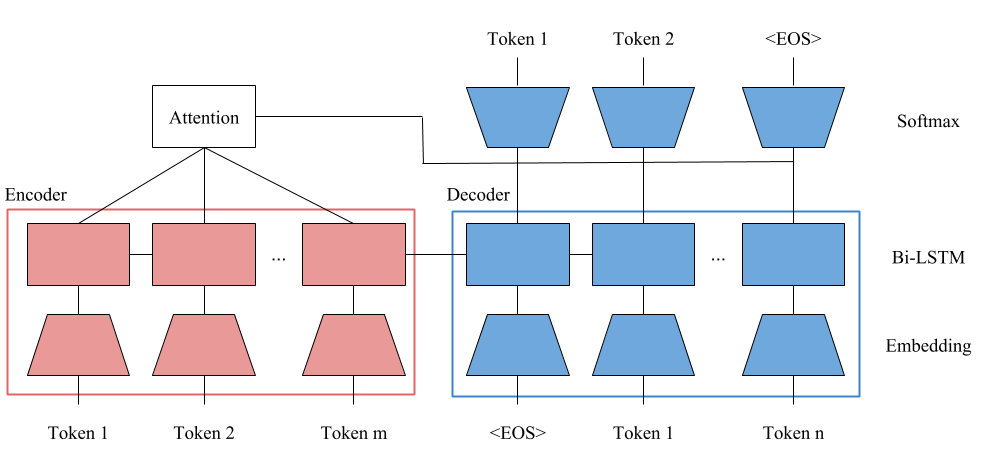}
    \caption{Seq-2-seq architecture with an attention-based copying mechanism}
\end{figure}

As depicted in Figure 1, the encoder takes the natural language utterances as a sequence of $m$ tokens. The first step is to lookup the embedding vector for the token and then to pass it through a bidirectional LSTM. The hidden states for the forward and backward RNNs are generated and the last hidden state is used as the initial hidden state for the decoder. At train time we run the natural language utterance through the encoder, initialize the decoder using the last hidden states, and then use teacher forcing to produce an output at each of the $n$ steps of the decoder while using the actual token for the next step.  The objective function we're optimizing at each Stochastic Gradient Descent (SGD) step is the standard negative log-likelihood computed for the softmax probability of the true token $y_t$ summed over the sequence. The per example loss is summed over all of the training examples $N$ to produce the total loss
\begin{equation}
-\sum_{i=1}^N \sum_{t=1}^{n} \log{p[\hat{y}_t=y_t]}  
\end{equation}

Three strategies are used for inducing a grammar over the original examples from each of the training datasets. The first strategy simply abstracts entities in the examples with their type. Assuming that each entity has a corresponding type the authors generate two rules. For instance the example with input \texttt{“what states border Texas?”} and output \verb|answer(NV, (state(V0), next_to(V0, NV), const(V0, stateid(texas))))| results in the following two recombinant rules:
\begin{itemize}
  \item
    \begin{lstlisting}
ROOT -> (what states border StateId?,
answer(NV, state(V0), next_to(V0, NV), const(V0, statid(StateId))
    \end{lstlisting}
  \item \verb|StateID -> (“texas”, texas)|
\end{itemize}

The second strategy abstracts whole phrases and adds one or two rules to the output grammar. The first of these two rules looks for matches between input and the output and replaces both of them with the type of the entity. Afterwards, if the entire output expression is of a particular types then a new rule is generated of the form \verb|type -> (input, output)|. Finally a $k$-concatenation strategy is used where $k >= 2$ is a parameter and simply creates a new rule that concatenates the sequence k times. The authors argue that the concatenation strategy has the effect of creating harder examples for the RNN to attend to, which has been shown to improve generalization.

\subsection {Co-occurrence augmentation}
We propose a novel augmentation strategy to be used in addition to the aforementioned strategies. This augmentation relies on the intuition that tokens which appear in similar contexts are interchangeable in both the source as well as the target sequences. For instance consider the following pair of natural language and logical form:
\begin{itemize}
  \item What is the capital of Alaska?
  \item
    \verb|_answer(NV,(_capital(V0),_loc(V0,NV),_const(V0,_stateid(alaska))))|
\end{itemize}
We can replace Alaska with any of the other 49 states in the US and get a valid pair. Our strategy for computing these equivalences is to look at source sentences with an equal number of tokens that only differ in a single token. These tokens are then linked together as we observe a minimum number of co-occurrences in the corpus. For example \verb|alaska| co-occurs with the following tokens: \verb|{alabama, arizona, arkansas, california, ...}|, which include over half of the US states.

This augmentation strategy also captures semantically similar tokens and allows for teaching these semantic invariances to the model. Consider the source sentence \verb|what is the highest point in ohio|. An example of an augmented semtence based on this sentence is \verb|what is the highest elevation in oklahoma|.

It should be noted that the co-occurrence strategy fails in some cases. For example the source sentence "\verb|list the states ?|" may produce the augmented sentence "\verb|list the california ?|". As the tokens \verb|states| and \verb|california| have been linked together due to co-occurrence in similar contexts. We have allowed for these cases to be produced so long as the generated logical form is valid. Our hypothesis is that the benefit of having the model generalize better by seeing correctly augmented examples outweigh the cost of producing semantically incorrect pairs.

\subsection{Pre-train and fine-tune} 
There's diminishing returns associated with data recombination in the original formulation. In particular, overusing data recombination can come at the cost of learning the actual task. We hypothesize that pre-training using recombinant data and subsequently fine-tuning the actual task can overcome this issue.

This framework is inspired by recent success stories such as BERT \cite{devlin2018bert}. To validate this hypothesis we start by simply breaking the original paper's single-stage training approach to the aforementioned two-stage approach. However, we only use recombinant examples for the pre-train stage and use the original, non-augmented training examples for fine-tuning. This framework allows for a high level of flexibility on the data, augmentation strategies, and objective functions that can be used for each stage as summarized in Table 1. For instance we can leverage unlabeled data and train the encoder and decoder as independent Language Models (LMs) as in \cite{ramachandran2016unsupervised}. In addition, we can use less precise recombinant strategies such as the co-occurrence strategy that would've been detrimental in the previous formulation.

\begin{table}[H]
  \caption{Data, augmentation, and objective functions used in the two stages of the proposed framework}
  \label{sample-table2}
  \centering
  \begin{tabular}{lll}
    \toprule
    \cmidrule(r){1-2}
                        & Pre-train                           & Fine-tune      \\
    \midrule
Leverage unlabeled data & Yes                                 & No             \\
Leverage other tasks    & Yes                                 & No             \\
Data recombination used & Yes                                 & No 
  \end{tabular}
\end{table}
The baseline for this work is the parsing and token accuracy reported for three standard semantic parsing datasets in \cite{jia2016data}. Parsing accuracy is defined as the proportion of the predicted logical forms that exactly match the true logical form \cite{liang2013learning}. It is possible that the same question could be captured by multiple valid logical forms and therefore a strict string match may be overly strict. Token accuracy is the proportion of the tokens in the true logical form that were present in the predicted sequence. We use parsing accuracy as a secondary metric as the model can learn to game it by producing more tokens. However, token accuracy is a useful metric for models that results in a similar parsing accuracy as well as a proxy for equally valid logical forms.

The original contribution of this work is the co-occurrence strategy as well as the proposed framework for pre-training and later fine-tuning without the recombinant strategies resulting in catastorphic interference or forgetting \cite{goodfellow2013empirical}. The code for the original paper was written in Theano and the code along with a number of utilities for preparing and evaluating the datasets and the experiments are all available in \url{https://worksheets.codalab.org/worksheets/0x50757a37779b485f89012e4ba03b6f4f/}. We have re-implemented the architecture and the two-stage framework in PyTorch and have used these utilities for preparation and evaluation. We have also used the OpenNMT library for experimenting with a variety of architectures such as Transformers. This library is available at \url{http://opennmt.net/OpenNMT-py/}.



\section{Experiments}

\subsection {Data}
We used the GeoQuery dataset which is available at \url{http://www.cs.utexas.edu/users/ml/geo.html}. This dataset consists of a set of questions about US geography facts and the corresponding Prolog query which represents the logical form. We use 600 examples for training and 280 examples for testing. The task is to produce the logical form given the question.

\subsection{Evaluation method}
Parsing accuracy (described in an earlier section) is used for evaluation.

\subsection{Details}
For the first experiment we explored the co-occurrence augmentation strategy as well as the two-stage training approach. Experiments are conducted with an encoder-decoder architecture with a bidirectional LSTM encoder with 256 hidden units and an LSTM decoder with the same number of hidden units. We have used a similar attention copy mechanism that was employed in \cite{jia2016data}, a learning rate of 0.001, word embedding size of 64, and 1,000 epochs of training. The original implementation in Theano used Stochastic Gradient Descent (SGD). We are using an Adam optimizer and using mini-batches of size 256 for training. Training the model in mini-batches resulted in significant speedups and allowed us to conduct many more experiments. Finally we have appended the training data with an equal number of augmented examples in the cases that augmentation was used.

After establishing the best combination of augmentation strategies we conducted hyper-parameter search to explore the effects of the embedding size, RNN hidden size, and the learning rate on the parsing accuracy.

Each experiment took an average of about 30 minutes to run on an AWS EC2 instance with 64 cores and 256GB of memory. We ran up to 32 experiments in parallel.

\subsection{Results}
Table 2 summarizes the sequence and token accuracy for the explored augmentation and pre-training strategies.

\begin{table}[H]
\centering
\caption{Augmentation and pre-training experiments}
\begin{tabular}{@{}llll@{}}
\toprule
Augmentation                           & Pre-train & Sequence accuracy & Token Accuracy \\ \midrule
nesting, entity, concat                & Yes       & \textbf{74.3}    & 87.8          \\
nesting, entity, concat                & No        & 73.6             & \textbf{88.2} \\
nesting, entity, concat, co-occurrence & Yes       & 66.1             & 87.3          \\
-                                      & Yes       & 62.1             & 81.6          \\
-                                      & No        & 58.6             & 81.6          \\
nesting, entity, concat, co-occurrence & No        & 56.1             & 81.7          \\
co-occurrence                          & Yes       & 55.4             & 79.8          \\
co-occurrence                          & No       & 51.7             & 80.7          \\ \bottomrule
\end{tabular}
\end{table}

The results suggest that pre-training can be marginally beneficial to sequence accuracy. However, the co-occurrence augmentation is leading to poorer results for this combination of hyper-parameters. One possible explanation is that this lower precision augmentation, by not enforcing a tight coupling between source and target sequences, is leading the model astray by producing semantically incorrect pairs that draw the wrong association between incompatible pairs which just happened to co-occur. We will see examples of this in the upcoming analysis section.

To gain more insight into the importance of the hyper-parameters we conducted search over the learning rate, word embedding size, and RNN hidden size. The results are summarized in Table 3.

\begin{table}[H]
\centering
\caption{Hyper-parameter search for best augmentation and pre-train combination}
\begin{tabular}{llrrr}
\toprule
      & Word embedding size &   32  &   64  &   128 \\
Learning rate & RNN hidden size &       &       &       \\
\midrule
0.001 & 64  &   0.0 &   2.1 &  11.1 \\
      & 128 &  24.6 &  32.1 &  63.2 \\
      & 256 &  70.4 &  \textbf{74.3} &  71.4 \\
0.010 & 64  &  40.7 &  45.4 &  36.1 \\
      & 128 &  68.9 &  68.2 &  59.6 \\
      & 256 &  66.1 &  55.0 &  69.6 \\
\bottomrule
\end{tabular}
\end{table}
The hyper-parameter search results suggest that a larger hidden size for the recurrent units is more beneficial while increasing the size of the embedding size has diminishing returns. This makes intuitive sense as the vocabulary size for this problem is very small and increasing the embedding size too much results in overparameterization.

We briefly experimented with using Transformers \cite{DBLP:journals/corr/VaswaniSPUJGKP17} for both the encoder and the decoder but did not manage to reproduce the same level of parsing accuracy and abandoned that line of investigation.

\section{Analysis}
In this section we analyze the best-performing model from the previous section. This model uses pre-training and employs all augmentation strategies except for co-occurrence. The model is trained with learning rate of 0.001, word embedding size of 64, and RNN hidden size of 256.

A useful metric for comparing the predicted and true sequences is the Intersection over Union (IoU) of the unique tokens in the two sequences:
\begin{equation}
\text{IoU}=\frac{\text{True} \cap \text{Predicted}}{\text{True} \cup \text{Predicted}}
\end{equation}
where $\text{True}$ is the set of non-punctuation tokens in the true logical form and $\text{Predicted}$ is the same set in the predicted sequence.

This metric focuses on how well the system is producing the correct tokens and disregards the order of the produced tokens and the correct nesting. For a correct parsing we get a value of 1, however an incorrect parsing can still get an \text{IoU} of 1 when the order or nesting is incorrect. Figure 2 shows the distribution of this metric for the incorrect results.

\begin{figure}[H]
    \centering
        \includegraphics[width=0.6\linewidth]{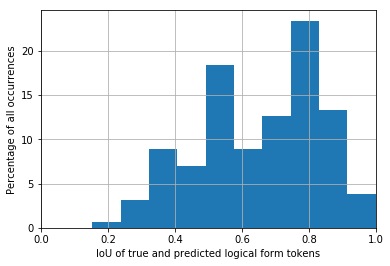}
    \caption{Distribution of the error attributable to the predicted tokens}
\end{figure}

In the vast majority of the cases the model is not even producing the correct tokens. Most of these are cases where the model is confusing similar predicates such as cities, states, mountains, or rivers. For example instead of generating \verb|_stateid(utah)| the model generates \verb|_cityid(utah)|. These mistakes results in the generation of incorrect logical forms.

A smaller proportion of the errors made by the model can be attributed to the logical form complexity. Figure 3 depicts the parsing accuracy against the number of open parentheses used in the true logical form as a measure of the logical form complexity.
\begin{figure}[H]
    \centering
        \includegraphics[width=0.6\linewidth]{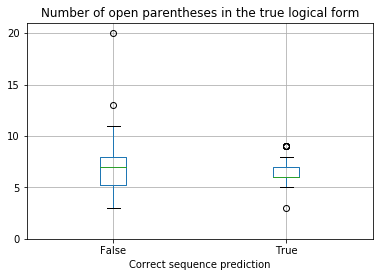}
    \caption{Effect of the logical form complexity on parsing accuracy}
\end{figure}
Logical forms with a larger number of opening parentheses tend to be more complex and have deeper nesting which make them harder to predict correctly. We can see that the model has not been able to correctly produce a logical form with more than 10 opening parentheses. To gain more insight into this process let's visualize the attention layer for a correctly predicted example in Figure 4.

\begin{figure}[H]
    \centering
        \includegraphics[width=0.6\linewidth]{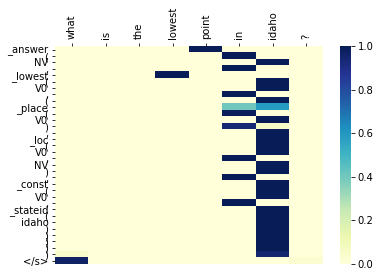}
    \caption{Attention weights for a correctly predicted example}
\end{figure}

We can see that the majority of the attention tends to be focused on a small subset of tokens in the source sentence. This lack of alignment makes it difficult for the model to switch attention between tokens in deeply nested logical forms as most of the instances in the training example do not take that form and look more similar to the example in Figure 4.

\section{Future work}
We showed that the two-stage framework proposed in this work can improve the parsing accuracy for a semantic parsing task. We need to extend the experiments conducted in this work to more datasets in order to establish a more definitive answer to whether these gains are persistent and significant beyond the GeoQuery dataset. We also need to explore other strategies for pre-training. In this work we used the same objective function for pre-training and fine-tuning. However, we will experiment with different strategies such as pre-training the decoder and encoder separately as language models. Especially for the decoder this can have the effect of teaching the model to produce valid logical forms using a large corpus.

The co-occurrence strategy proposed in this work did not prove to be promising in its current form. As we hypothesized earlier this may be due to the fact that co-occurring tokens may be linked together by appearing in similar contexts. We strategy to overcome this problem is to restrict the augmentation to cases that do not affect any of the predicates in the logical form. This higher precision strategy may reduce the production of semantically incorrect natural language utterances while teaching the model about semantic invariances.



\end{document}